\documentclass{article}

    \usepackage[nonatbib,final]{neurips_2020}

\usepackage[utf8]{inputenc} 
\usepackage[T1]{fontenc}    
\usepackage{hyperref}       
\usepackage{url}            
\usepackage{booktabs}       
\usepackage{amsfonts}       
\usepackage{nicefrac}       
\usepackage{microtype}      
\usepackage{xcolor}
\usepackage{graphicx}
\usepackage{xspace}
\usepackage{float}

\title{OGNet: Towards a Global Oil and Gas \\ Infrastructure Database using\\ Deep Learning on Remotely Sensed Imagery}

\author{%
    Hao Sheng\thanks{Equal contribution.} ~$^1$, Jeremy Irvin\footnotemark[1]~~$^1$\\ 
    \textbf{Sasankh Munukutla$^1$, Shawn Zhang$^1$,  Christopher Cross$^1$} \\ 
    \textbf{Kyle Story$^2$, Rose Rustowicz$^2$,  Cooper Elsworth$^2$, Zutao Yang$^1$} \\ 
    \textbf{Mark Omara$^3$, Ritesh Gautam$^3$, Robert B. Jackson$^1$, Andrew Y. Ng$^1$ }\\
    $^1$Stanford University, $^2$Descartes Labs, $^3$Environmental Defense Fund\\
    \small{\texttt{\{haosheng,jirvin16,sasankh,szhang22,chrisglc\}}\texttt{@cs.stanford.edu}} \\
    \small{\texttt{\{kyle,rose,cooper\}}\texttt{@descarteslabs.com}},
    \small{\texttt{\{momara,rgautam\}}\texttt{@edf.org}} \\
    \small{\texttt{rob.jackson@stanford.edu}},   \small{\texttt{ang@cs.stanford.edu}}
}

\begin{document}

\maketitle

\newif\ifshowcomments
\showcommentsfalse
\ifshowcomments
\newcommand\jeremy[1]{\textcolor{red}{[jeremy: #1]}}
\newcommand\hao[1]{\textcolor{blue}{[hao: #1]}}
\newcommand\sasankh[1]{\textcolor{green}{[sasankh: #1]}}
\newcommand\shawn[1]{\textcolor{purple}{[shawn: #1]}}
\newcommand\chris[1]{\textcolor{orange}{[chris: #1]}}
\else
\newcommand\jeremy[1]{}
\newcommand\hao[1]{}
\newcommand\sasankh[1]{}
\newcommand\shawn[1]{}
\newcommand\chris[1]{}
\fi

\newcommand{\model}{DefoRestNet\xspace}

\begin{abstract}
At least a quarter of the warming that the Earth is experiencing today is due to anthropogenic methane emissions.  There are multiple satellites in orbit and planned for launch in the next few years which can detect and quantify these emissions; however, to attribute methane emissions to their sources on the ground, a comprehensive database of the locations and characteristics of emission sources worldwide is essential.  In this work, we develop deep learning algorithms that leverage freely available high-resolution aerial imagery to automatically detect oil and gas infrastructure, one of the largest contributors to global methane emissions. We use the best algorithm, which we call OGNet, together with expert review to identify the locations of oil refineries and petroleum terminals in the U.S. We show that OGNet detects many facilities which are not present in four standard public datasets of oil and gas infrastructure. All detected facilities are associated with characteristics known to contribute to methane emissions, including the infrastructure type and the number of storage tanks. The data curated and produced in this study is freely available at \url{http://stanfordmlgroup.github.io/projects/ognet}.
\end{abstract}

\section{Introduction}
Methane is the second-largest contributor to climate warming after carbon dioxide and accounts for at least one-quarter of present-day warming \cite{myhre2013anthropogenic}. Around a third of current global anthropogenic methane emissions arise from the fossil fuel sector, and reducing these emissions is critical to mitigate further warming in the near-term \cite{maasakkers2019global,hmiel2020preindustrial,shoemaker2013role, jackson2020increasing}. In the last decade, quantifying methane emissions from the oil and gas (O\&G) sector has become a significant area of research and interest to both governments and the O\&G industry \cite{alvarez2018assessment,epa2004gmi,zhang2020quantifying,scarpelli2020global}. Recent progress in quantifying methane emissions from the O\&G sector has been largely due to advancements in satellite-based observations \cite{kort2014four,pandey2019satellite,de2020daily,zhang2020quantifying,schneising2020remote}. Currently, multiple global mapping satellite missions (SCIAMACHY, GOSAT, TROPOMI) provide atmospheric methane concentrations data \cite{jacob2016satellite} and limited coverage satellites (GHGSat) can detect high emission rates from specific facilities \cite{varon2019satellite}. Furthermore, a new satellite (MethaneSAT) planned for launch in 2022 will provide publicly-available methane emissions data made up of diffuse emission fields and point sources to help drive action towards reducing emissions and for companies and countries to track their emission reduction targets \cite{wofsy2019methanesat,methanesat}.

To attribute methane emissions detected by these satellites to facilities on the ground, a comprehensive database of the locations and characteristics of O\&G infrastructure worldwide is critically required. There have been prior efforts to construct public datasets of O\&G facilities. The global O\&G infrastructure database (GOGI) was curated by developing machine learning models to parse the web for publicly-listed instances of infrastructure, but this process leads to data gaps due to limitations with searching for records published online \cite{rose2018global}. Other national programs which host O\&G infrastructure data include the Homeland Infrastructure Foundation-Level Data (HIFLD) by the Department of Homeland Security \cite{hifld}, the U.S Energy Mapping System from the Energy Information Administration (EIA) \cite{eia}, and the Greenhouse Gas Reporting Program (GHGRP) by the EPA \cite{ghgrp}. The data sources for these programs include required reporting by facility operators and publicly available data, which may be outdated or have incomplete coverage \cite{national2018improving}.

The recent unprecedented availability of high-resolution satellite and aerial imagery, together with advancements in deep learning methods, offer a powerful opportunity to create high-quality, granular, and large-scale databases of global infrastructure \cite{karpatne2018machine,janowicz2020geoai}. Deep learning techniques have been increasingly utilized for automatically detecting objects in remotely sensed imagery, from building footprint detection \cite{li2019semantic} and urban land use classification \cite{zhang2018urban} to energy infrastructure classification, including solar photovoltaics \cite{yu2018deepsolar} and wind turbines \cite{zhoudeepwind}. Only recently have these models been deployed on high-resolution imagery to construct large-scale databases \cite{yu2018deepsolar,hou2019solarnet}.

In this work, we develop deep learning models to automatically detect O\&G infrastructure in freely available aerial imagery. We deploy the best models in the full continental U.S. to map oil refineries and petroleum terminals and attribute each facility with characteristics that are known to contribute to methane emissions. To our knowledge, this is the first study to leverage high-resolution imagery to map O\&G facilities in the continental U.S.  All data curated and generated in this study is available at \url{http://stanfordmlgroup.github.io/projects/ognet}.

\section{Methods}

\setlength{\tabcolsep}{2pt}
\begin{figure*}[t!]
  \centering
  \begin{tabular}{ccc}
    \includegraphics[scale=0.13]{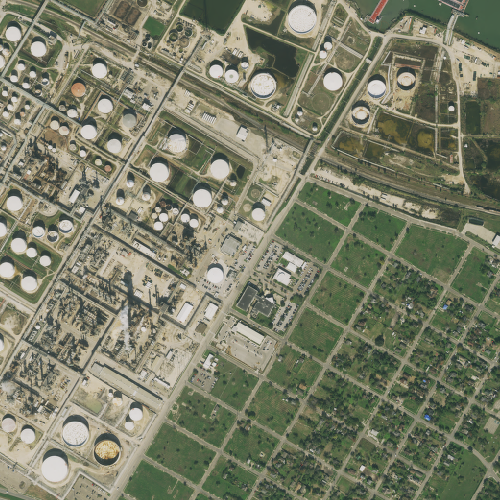} & \includegraphics[scale=0.13]{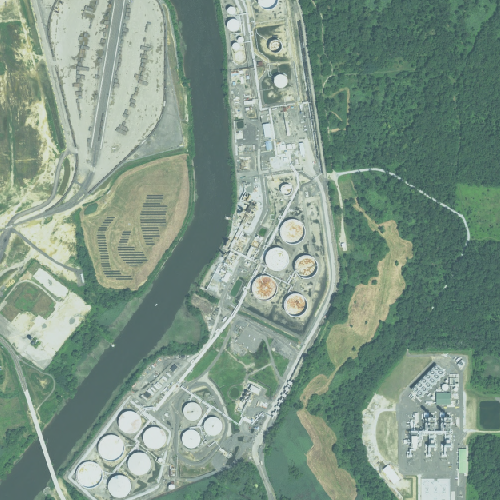} & \includegraphics[scale=0.13]{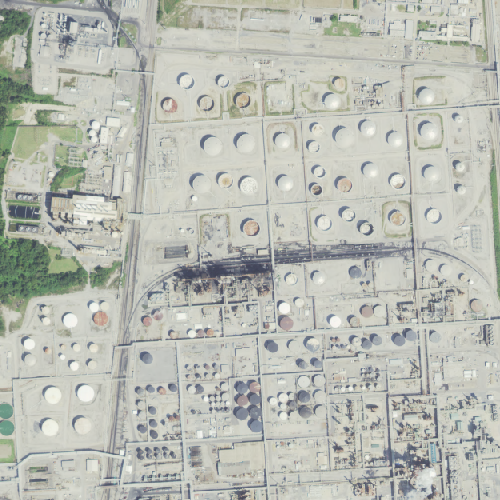}  \\[0.1cm]
    \includegraphics[scale=0.13]{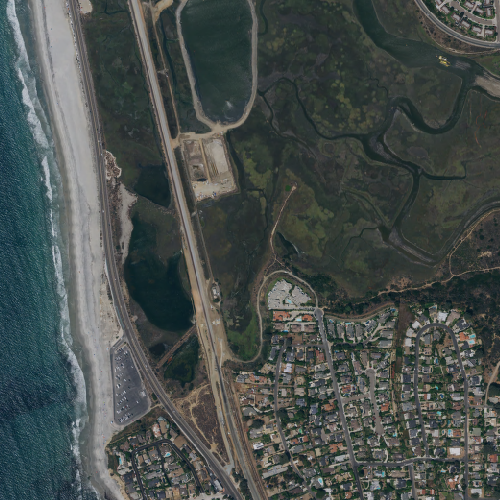} & \includegraphics[scale=0.13]{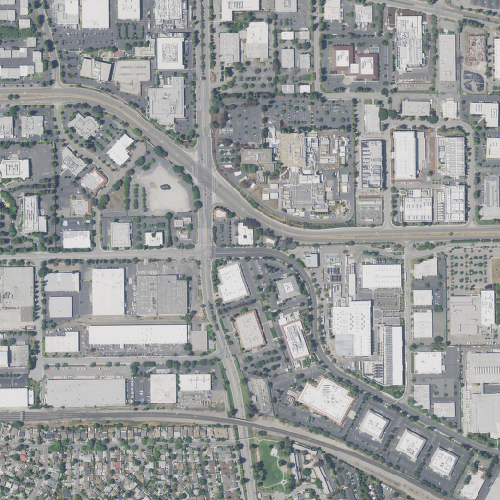} & \includegraphics[scale=0.13]{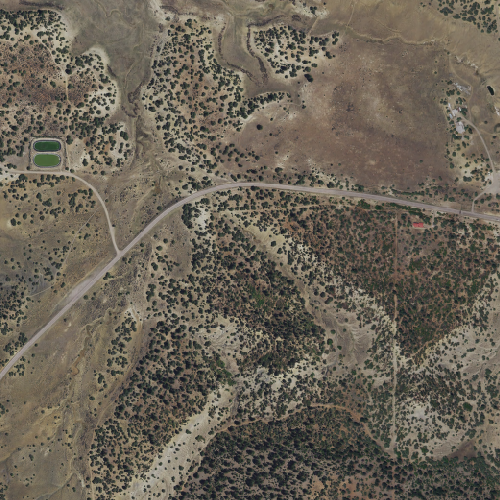}
\end{tabular}
  \caption{Example NAIP imagery of positive examples (top row) and negative examples (bottom row) in the dataset used for model development. The positive images capture oil refineries, characterized by a large footprint and consistent features like large round storage tanks. The negative images consist of randomly sampled locations as well as objects and landscapes which look similar to oil refineries.}
  \label{fig:data}
\end{figure*}

\paragraph{Task}
We train deep learning models to perform a canonical binary classification task, namely identifying whether an aerial image contains an oil refinery. We focus on oil refineries as they have large footprints and consistent features, which make them feasible to detect in aerial imagery, but we demonstrate that the deep learning models can also detect other O\&G facilities.

\paragraph{Dataset Instances} To develop and validate the models, we use 149 locations of oil refineries in the U.S. obtained from the Enverus Drillinginfo database, a commercial database containing point locations of various types of O\&G infrastructure \cite{enverus}. We additionally curate a large number of negative examples to include in the dataset. We first include a random sample of locations in the U.S. that capture a variety of landscapes. However, this set of negatives does not include facilities that may appear similar to oil refineries or landscapes that may commonly appear near oil refineries. In order to address this, we sample locations containing visually similar objects and landscapes (difficult negatives) using an open-source GeoVisual search tool \cite{keisler2019visual}. The query locations used in the search were obtained by finding objects and landscapes near the locations of the oil refineries in the dataset (see Appendix for details). The full dataset contains 7,066 examples in total and was split into a training set to learn model parameters (127 positive examples, 5,525 negative examples), a validation set to select model hyperparameters (13 positive examples, 693 negative examples), and a test set to evaluate the performance of the best model (9 positive examples, 697 negative examples).

\paragraph{Aerial Imagery}
We use aerial imagery between 2015 and 2019 from the National Agriculture Imagery Program (NAIP) \cite{naip}, which captures the continental U.S. at a minimum of 1m resolution. We construct images of size 500 x 500 pixels downsampled to a resolution of 2.5m. We found that this area was sufficient to capture a large portion of the facility in most of the cases while also balancing memory usage and resolution for accurate detection. The images are mosaics of the most recent captures of the location and do not suffer from cloud cover or haze because images were acquired aerially on days with low cloud cover. The images of positive examples are centered around the coordinates associated with the oil refineries. Example images are shown in Figure~\ref{fig:data}.


\begin{figure*}[t!]
  \centering
  \includegraphics[scale=0.19]{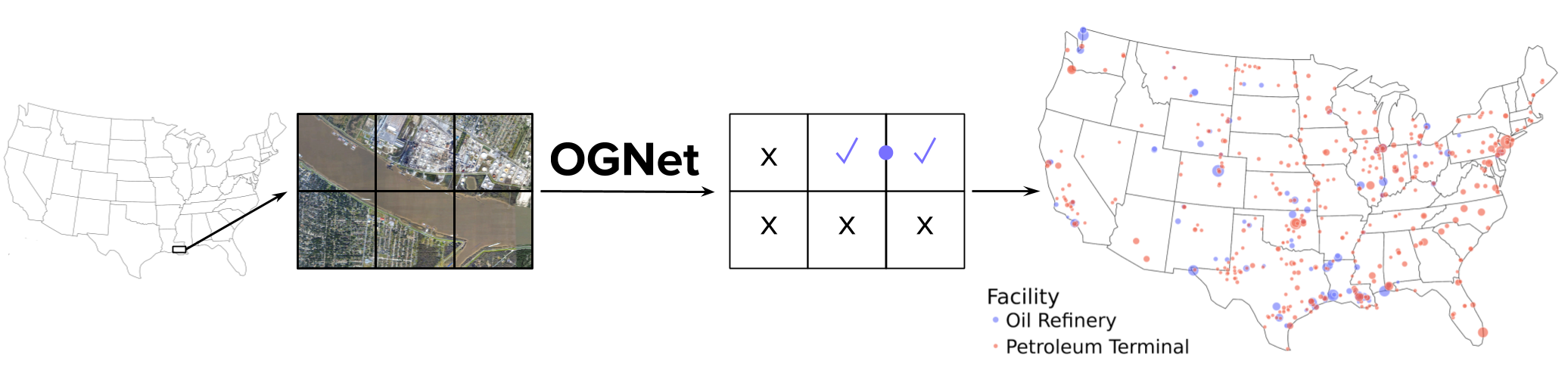}
  \caption{
    OGNet classifies NAIP imagery for the presence of oil and gas infrastructure. We deploy OGNet on the entire U.S. to identify the locations of oil refineries and petroleum terminals. The radius of the circles is proportional to the number of storage tanks at each facility.
  }
  \label{fig:model}
\end{figure*}

\paragraph{OGNet Development and Deployment}
The deep learning model training procedure and architectures that were explored are provided in the Appendix. 
We call our best model OGNet. We convert predicted probabilities to binary predictions using the threshold that achieves the highest precision (0.81) subject to a recall of 1.0 on the validation set. 
To identify infrastructure in the continental U.S. using OGNet, we partition the region into equal 500 x 500 pixel tiles at 2.5m resolution leading to 5,082,722 tiles in total and run each of the tiles through OGNet. 
Any tile which was assigned a positive prediction is greedily merged with any positively classified adjacent tiles, and the mean of the centroids of each of the tiles within the merged group is used as the detected location (see Figure~\ref{fig:model}). 

\paragraph{Human Review}
OGNet detections in the continental U.S. were manually reviewed, and all false positive detections were removed.
We discovered that OGNet was able to detect O\&G facilities other than oil refineries, so the remaining facilities were classified as oil refineries or petroleum terminals (including liquefied natural gas terminals and crude oil terminals). During this process, we also identified the number of storage tanks at each facility. An expert interpreter designed the annotation procedure (see Appendix) and verified the detected facilities and their attributed characteristics.

\section{Results}

\paragraph{OGNet Test Set Performance}
The best performing model, which we call OGNet, was a 121-layer DenseNet model. At its operating point, it achieved an accuracy of 0.996, a precision of 0.75, a recall of 1.0, and an F1 of 0.857 on the test set. The false positives all contained characteristic features of oil refineries but were other facilities, including a wastewater treatment plant (see Appendix).

\paragraph{OGNet Deployment Results}
OGNet detected a total of 1,902 instances of potential facilities in the U.S. after filtering two regions that contained many false positive detections. After manual review, 114 of the detected facilities were classified as oil refineries, and 336 detected facilities were classified as petroleum terminals. The remaining detections were false positives, and typical examples that OGNet assigned high probability were grain elevators and water treatment and chemical plants. To interpret model predictions, we use class activation maps (see Appendix for details and examples).

\paragraph{Comparison with Public Datasets}
We compare the list of manually verified facilities detected by OGNet with four publicly available datasets, namely GOGI, GHGRP, HIFLD, and EIA. We combine the datasets and remove duplicate records by combining coordinates within 2 km of each other. To match the coverage of NAIP imagery, we additionally remove any samples outside the continental U.S. The final combined dataset has 147 oil refineries and 1,222 petroleum terminals in total.

For each facility in the combined dataset, we count the facility as covered if there is a facility detected by OGNet within 3 km, chosen to account for any perturbations in coordinates due to how centroid coordinates were determined in the detections. OGNet detected 73.5\% of the oil refineries and 23.9\% of the petroleum terminals in the combined dataset. Close to half of the ``missed'' oil refineries were due to inaccurate locations reported in the public datasets, with reported coordinates deviating more than 5 km away from the actual site. We also count the number of detected facilities that neither occur in the public datasets nor the training set. OGNet detected 6 oil refineries (including one abandoned facility) and 142 petroleum terminals that were not present in the public datasets (Table~\ref{table:dataset}).

\begin{table}[t!]
  \begin{center}
  {\fontfamily{cmss}\selectfont
  \resizebox{0.5\textwidth}{!}{
    \begin{tabular}[t]{l|c|c}
      \toprule
        & Oil Refinery &  Petroleum Terminal \\
      \midrule
      Total Detections & 114 & 336 \\
      \midrule
      Coverage of Benchmark Datasets & 73.5\% (108/147) & 23.9\% (292/1222) \\
      \midrule
      New Detections & 6  & 142 \\
      \midrule
      Example Image &\includegraphics[scale=0.085]{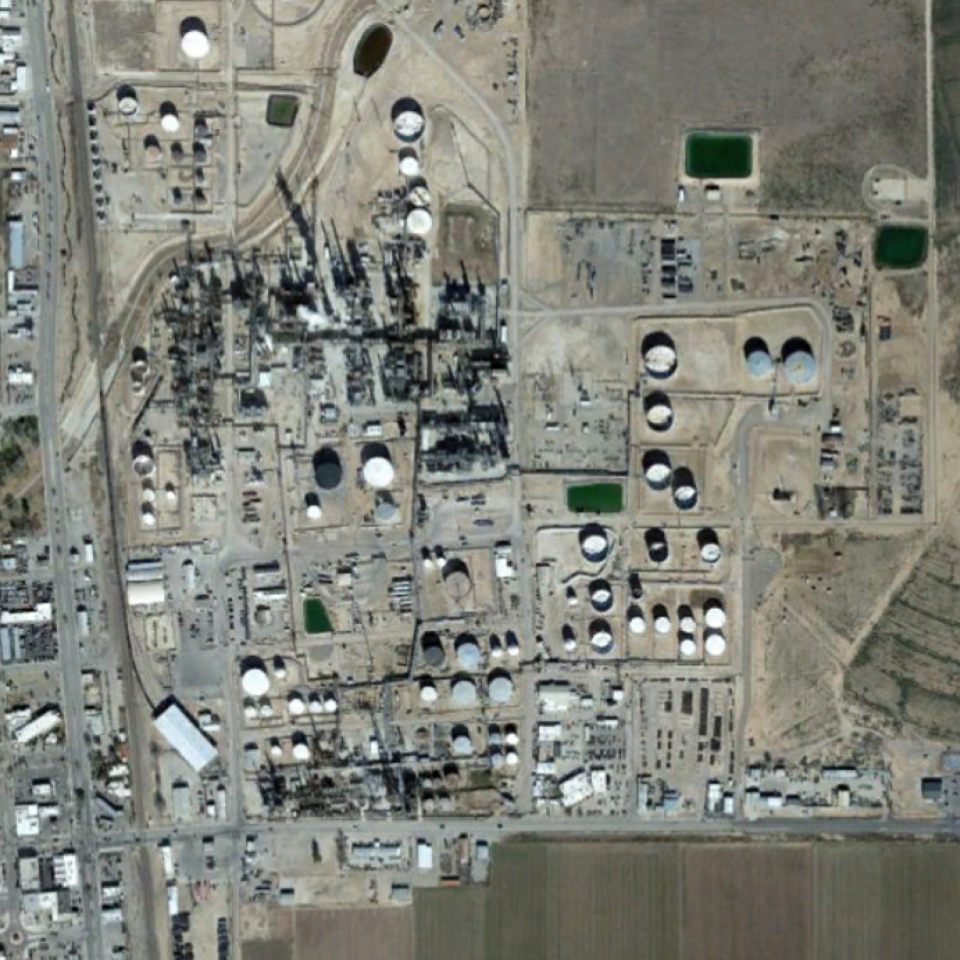} &\includegraphics[scale=0.085]{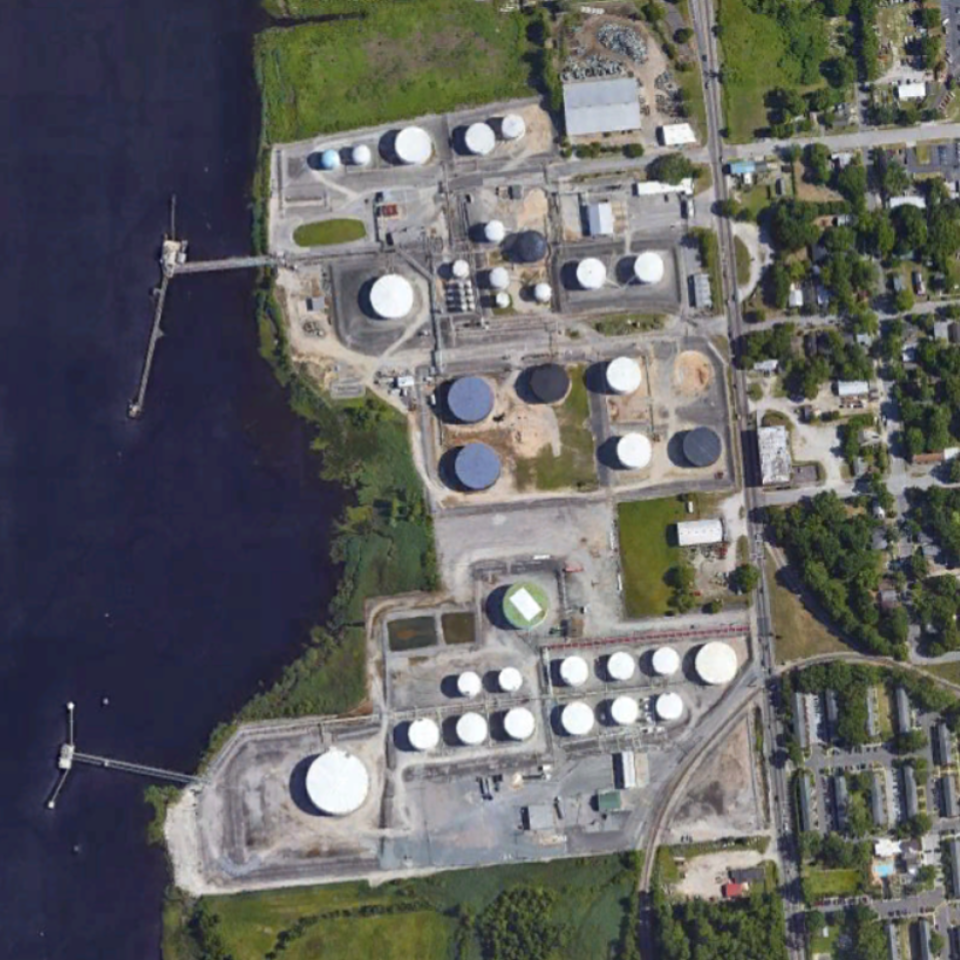} \\
      \bottomrule
    \end{tabular}
`    }
    }
  \end{center}
  \caption{Statistics of the OGNet detections and comparison to benchmark datasets. We compare detections to the union of four public databases of O\&G infrastructure, including GOGI \cite{rose2018global}, GHGRP \cite{ghgrp}, HIFLD \cite{hifld}, and EIA \cite{eia}. New detections do not include any instances in the training set.}
  \label{table:dataset}
\end{table} 

\section{Discussion}
Our results demonstrate the capability of deep learning models to detect large O\&G facilities in remotely sensed imagery. We believe that the development and deployment of these models can fill existing gaps in publicly reported data. While we demonstrate the feasibility of detecting oil refinery and petroleum terminals in the U.S., our approach can be extended to other O\&G production regions around the world using high-resolution imagery and thus has the potential to address critical data gaps, particularly in regions with a lack of data. We anticipate that the methods presented in this work can be extended to identify other types of O\&G infrastructure and combined with other sources of data to create a comprehensive, granular, and up-to-date global database of O\&G infrastructure.

We believe the development of this database will be important to the site-level quantification and attribution of greenhouse gas emissions, especially when combined with coarse measurements of emissions from other sources~\cite{wofsy2019methanesat,methanesat}. Our work is a first step towards building a comprehensive infrastructure inventory to support multiple methane-measuring satellite missions aiming to attribute emissions and monitor progress toward emissions abatement. Focusing on the O\&G sector is crucial as it is among the largest contributors to global anthropogenic methane emissions, and furthermore, there is evidence that methane emissions from fossil fuels have been significantly underestimated \cite{hmiel2020preindustrial}.

This study has important limitations. First, high-resolution imagery like NAIP is not widely publicly available worldwide. Future work should investigate the use of imagery with global coverage like Sentinel-2 10m optical imagery \cite{drusch2012sentinel}. Second, while the model dramatically reduces the necessary amount of manual review, manual review is still necessary, and the current approach does not leverage the new annotations during training. Human-in-the-loop machine learning techniques could be used to incorporate human feedback \cite{xin2018accelerating}. Third, a canonical classification approach may not be effective for O\&G infrastructure with smaller footprints like well pads and compressor stations. Object detection and semantic segmentation models may be more effective for localizing other O\&G infrastructure.




\small
\bibliography{bibliography}
\bibliographystyle{ieeetr}

\newpage
\appendix
\section*{Appendix}
\setcounter{table}{0}
\setcounter{figure}{0}
\renewcommand{\thetable}{A\arabic{table}}
\renewcommand{\thefigure}{A\arabic{figure}}

\subsection*{Examples}
\begin{figure*}[ht!]
  \centering
  \begin{tabular}{ccc}
    \includegraphics[scale=0.12]{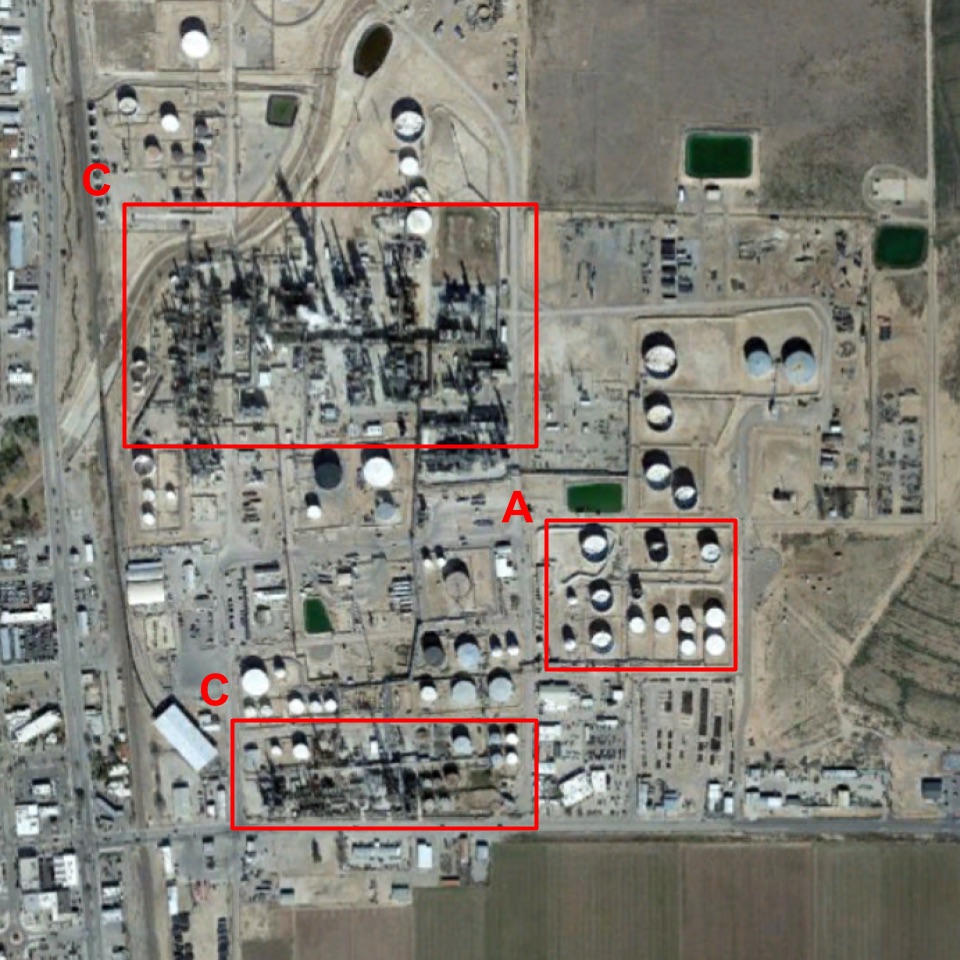} 
  & \includegraphics[scale=0.12]{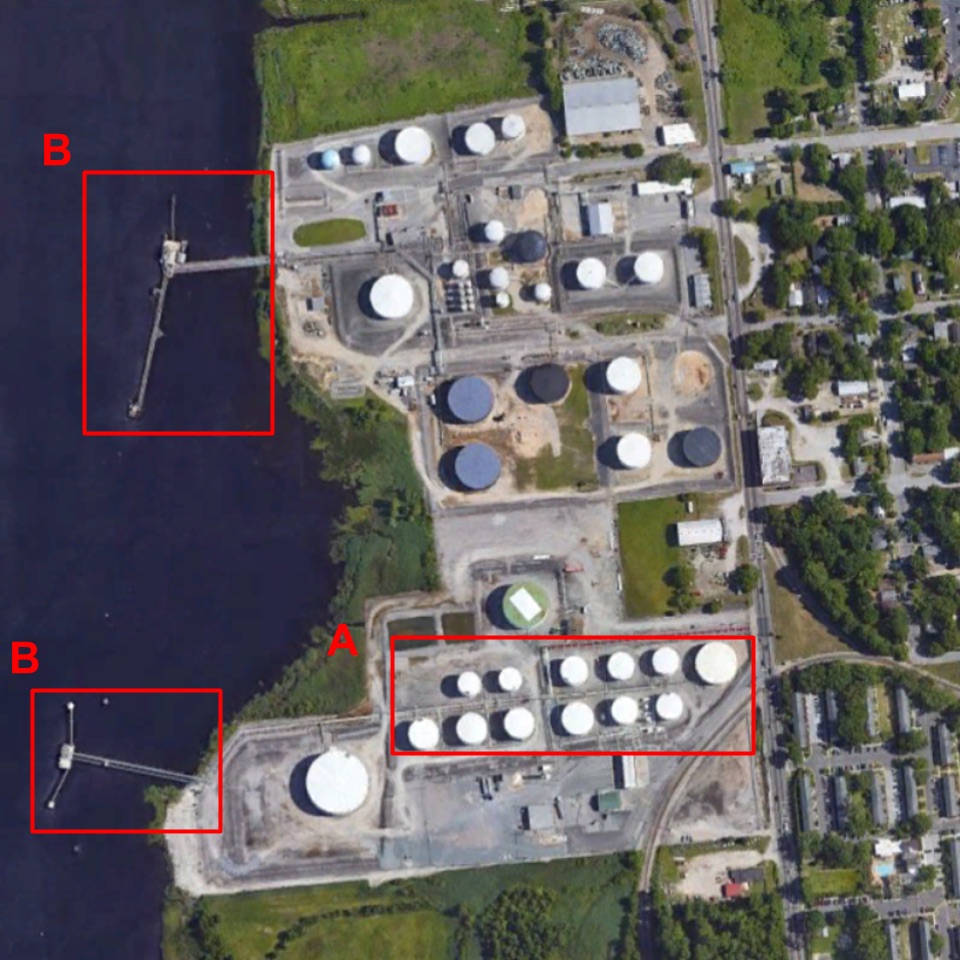} 
  & \includegraphics[scale=0.12]{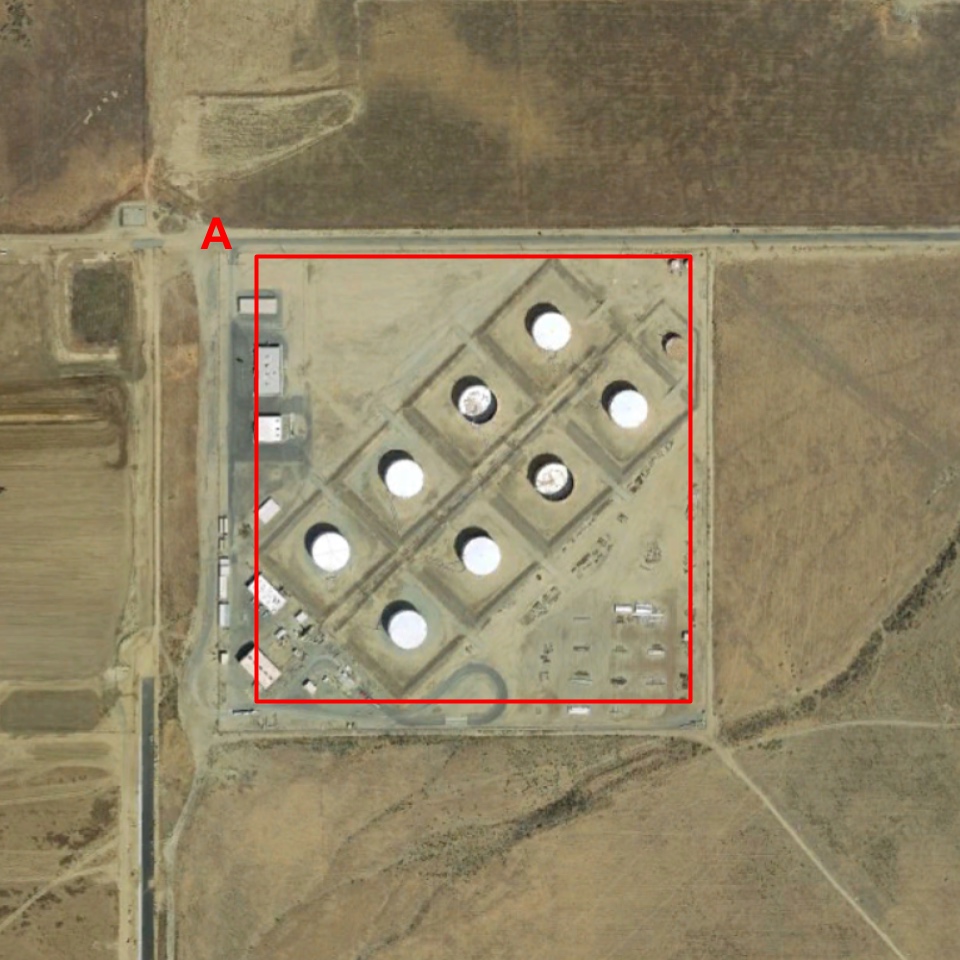}  
\end{tabular}
  \caption{Examples of features (in bounding boxes {\bf (A)(B)(C)}) associated with oil and gas infrastructure. {\bf (A)} {\it Storage containers and tank farms} tend to be common indicators for all types of infrastructure (both oil refineries and petroleum terminals). {\bf (B)} {\it Jetties and piers} tend to be exclusive to LNG terminals and occasionally coastal refineries. {\bf (C)} {\it Distillation units} are typically unique to refineries. The high resolution images shown here are obtained from Google Earth.}
  \label{fig:facilities}
\end{figure*}
\begin{figure*}[ht!]
  \centering
  \begin{tabular}{ccc}
    \includegraphics[scale=0.12]{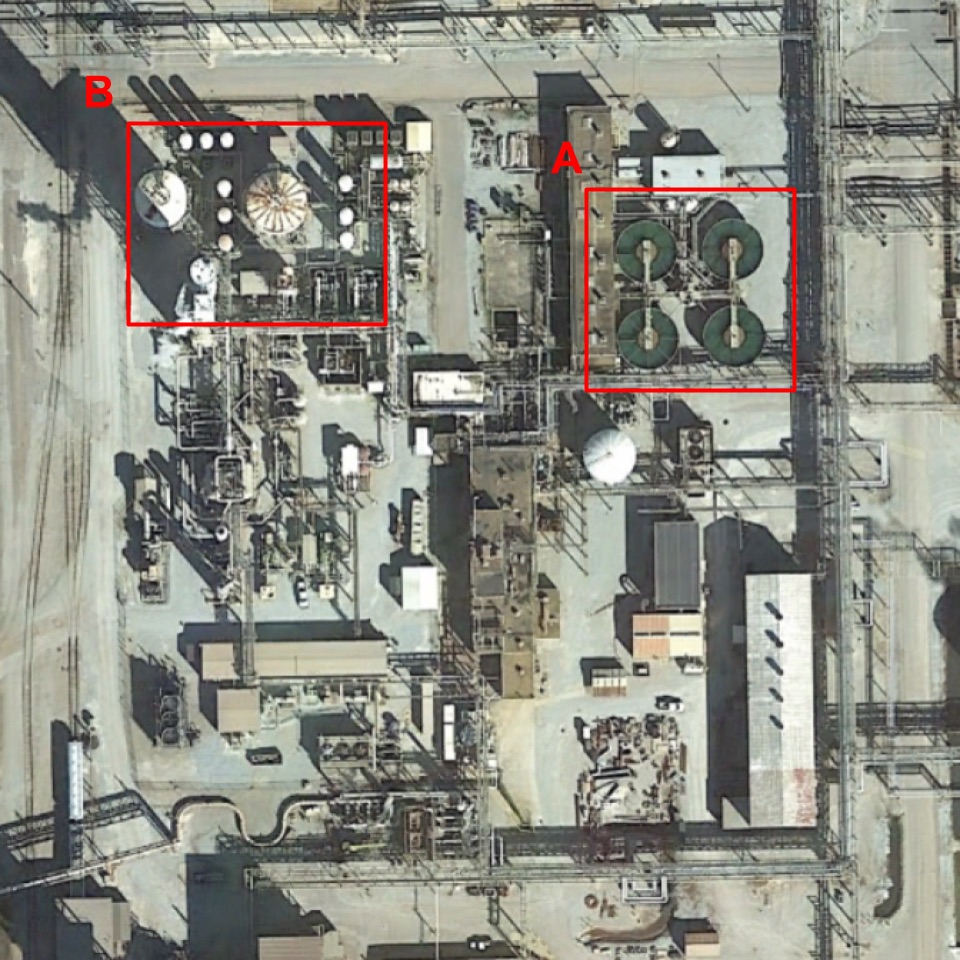} 
  & \includegraphics[scale=0.12]{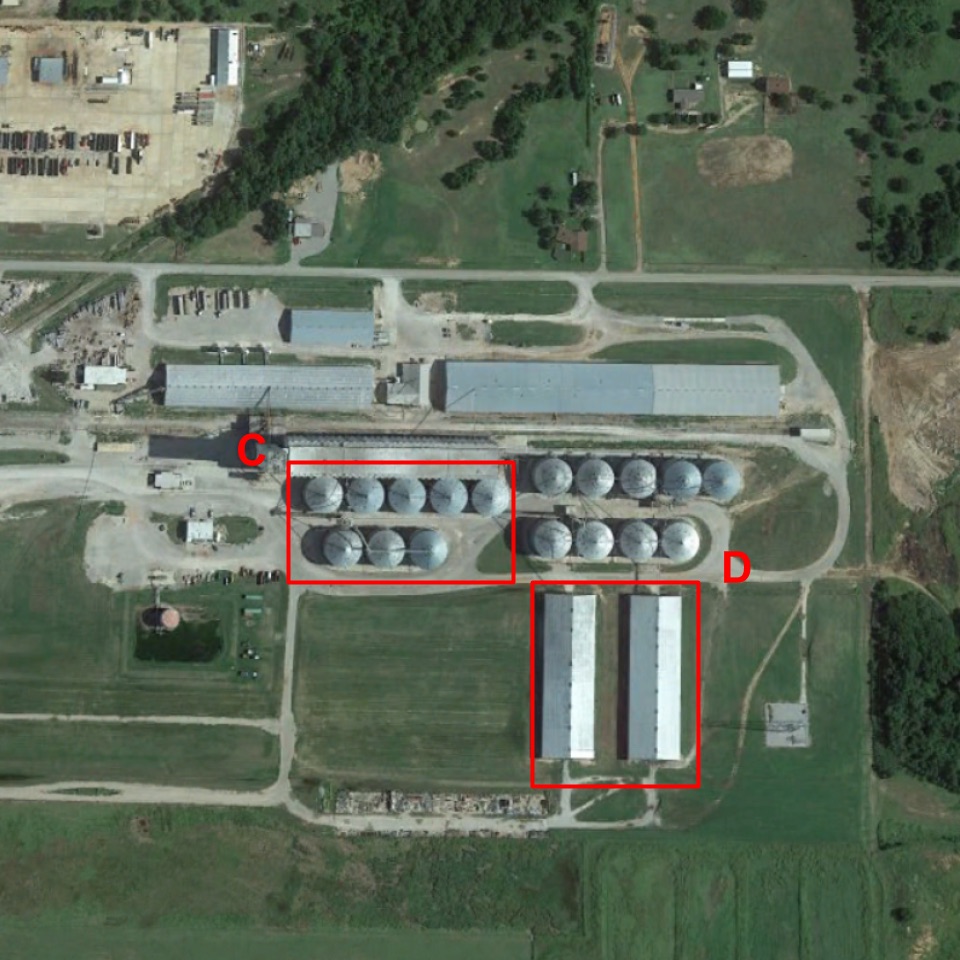} 
  & \includegraphics[scale=0.12]{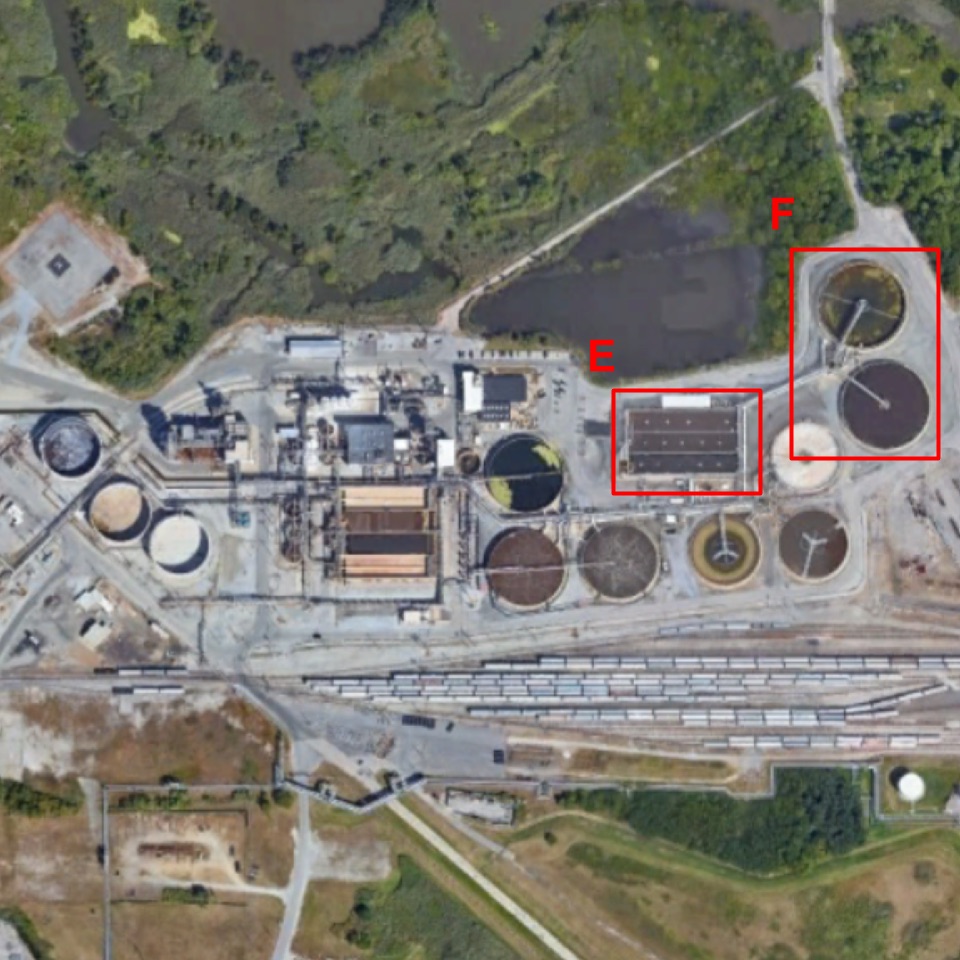}  
\end{tabular}
  \caption{Examples of features (in bounding boxes {\bf(A-F)}) associated with common facilities which mimic oil and gas infrastructure. (Left) Chemical plants are usually indicated by {\bf (A)} {\it clarifiers}, {\bf(B)} {\it small processing tanks}, and overall complex footprints similar to refineries. (Middle) Grain processing facilities are usually indicated by {\bf (C)} {\it grain bins} and {\bf (D)} {\it storage warehouses}. (Right) Wastewater treatment facilities are usually indicated by {\bf (E)} {\it sedimentation tanks} and {\bf (F)} {\it wastewater clarifiers}. The high resolution images shown here are obtained from Google Earth.}
  \label{fig:other_facilities}
\end{figure*}
\begin{figure*}[t!]
  \centering
  \begin{tabular}{ccc}
    \includegraphics[scale=0.23]{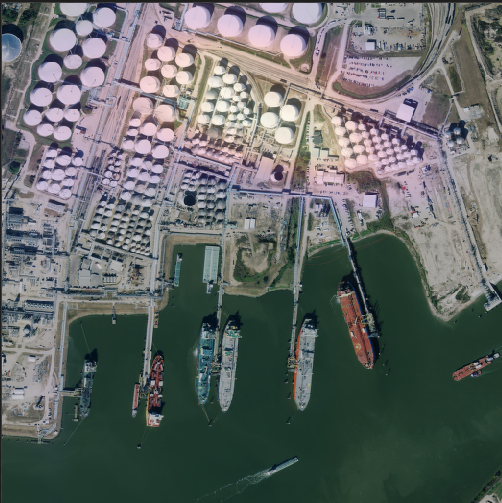} 
  & \includegraphics[scale=0.23]{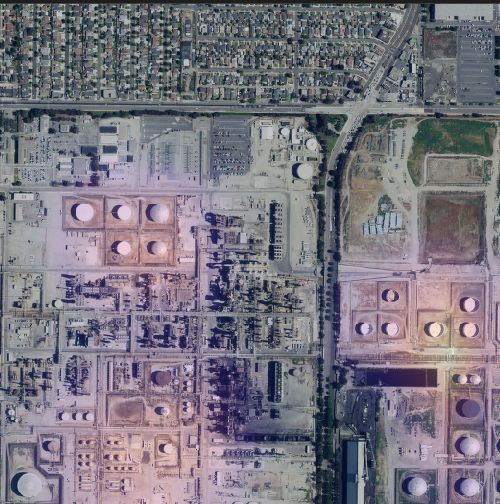} 
  & \includegraphics[scale=0.23]{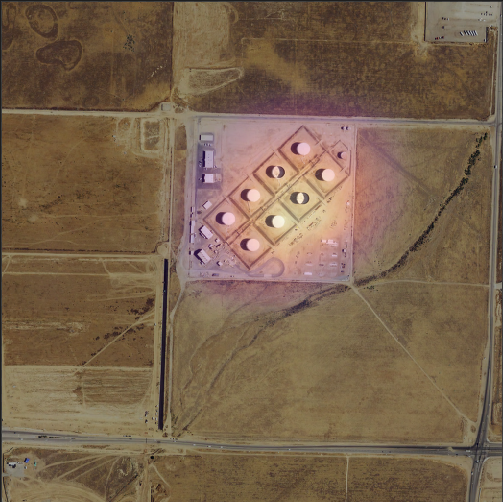}  
\end{tabular}
  \caption{Examples of Class Activation Maps (CAMs) on OGNet detections from deployment. CAMs are produced by OGNet and overlaid over the NAIP images to highlight features of the image which contribute most to its prediction. The CAMs shown here highlight indicative features like storage tanks and distillation units mentioned in Figure~\ref{fig:facilities}.}
  \label{fig:cams}
\end{figure*}

\subsection*{Difficult Negative Sampling} We sampled difficult negatives to include in the training, validation, and test sets in order to capture infrastructure and landscapes which may be misclassified by the model. We sampled these examples in multiple stages based on the performance of the model on the deployment set. First, we included locations with common detailed features similar to those of oil refineries including highly-populated urban landscapes, suburban landscapes, and townships with sparse shrubbery.
Second, we included more challenging examples that led to frequent false positives including well pads, arable land/crop fields, and dense forestry. Third, we included snowy regions that the model tended to conflate with the features of oil and gas infrastructure (e.g. tops of white storage tanks). We found that the snowy regions reduced sensitivity to refineries, so the final model did not use these images in the dataset.

\subsection*{Architecture and Training Procedure}  We experimented with ResNet \cite{he2016deep} and DenseNet \cite{huang2017densely} architectures with different numbers of layers. We applied a variety of augmentations during training including random vertical and horizontal flips, random affine transformations and random color jitter. We initialized the networks with weights from a model pre-trained on ImageNet \cite{deng2009imagenet} and normalized the images based on the mean and standard deviation of images in the ImageNet training set. The networks were trained end-to-end using a unweighted binary cross entropy function, Adam with standard parameters \cite{kingma2014adam}, and a learning rate of 0.0001. During training, we evaluated the network on the validation set after each epoch and save the checkpoint with the lowest validation loss.  We call the model which achieved the best performance on the validation set OGNet.

\subsection*{Deployment Run-time}
Constructing and downloading the mosaics used in deployment took 20 days on a single machine with multiprocessing, and generating the predictions on the images using a single NVIDIA TITAN-Xp GPU took 40 hours.

\subsection*{Annotation Procedure}
An expert on the oil and gas sector designed the manual annotation procedure for verifying and characterizing facilities detected by OGNet. Each detection was classified as negative, oil refinery, crude oil terminal, or liquefied natural gas terminals. The oil and gas facilities were identified using characteristic features visible from sub-meter resolution imagery provided by Google Maps and Google Earth Pro (Figure ~\ref{fig:facilities}). Other facilities with similar features were distinguished by identifying the presence of other objects that are not present in the oil and gas facilities (Figure ~\ref{fig:other_facilities}) and using place names in Google Maps. Finally, the number of big (internal and external) floating roof tanks was counted at each facility, verified using Google Earth's 3D building view.

\subsection*{Class Activation Maps}
We used class activation maps (CAMs) to interpret OGNet predictions \cite{zhou2016learning}. CAMs highlight the regions of the image which contribute to a positive classification. The CAM for an image was computed by taking a weighted average between the feature maps produced by OGNet for that image and the weights of the fully connected layer, followed by setting all negative values to zero. For an input image of size 500 x 500 pixels, the CAM was of size 15 x 15. We upsampled the CAM to size 500 x 500 and overlaid the image to highlight salient regions. Examples of class activation maps produced by OGNet are shown in Figure~\ref{fig:cams}.

\end{document}